\RequirePackage[bookmarksnumbered,unicode]{hyperref}
\documentclass[sigconf,natbib=false]{acmart}

\AtBeginDocument{%
  \providecommand\BibTeX{{%
    \normalfont B\kern-0.5em{\scshape i\kern-0.25em b}\kern-0.8em\TeX}}}

\acmDOI{10.1145/3589132.3625638}

\acmPrice{15.00}
\acmISBN{978-1-4503-XXXX-X/18/06}

\copyrightyear{2023}
\acmYear{2023}
\setcopyright{rightsretained}
\acmConference[SIGSPATIAL '23]{The 31st ACM International Conference on
Advances in Geographic Information Systems}{November 13--16,
2023}{Hamburg, Germany}
\acmBooktitle{The 31st ACM International Conference on Advances in
Geographic Information Systems (SIGSPATIAL '23), November 13--16, 2023,
Hamburg, Germany}\acmDOI{10.1145/3589132.3625638}
\acmISBN{979-8-4007-0168-9/23/11}
\RequirePackage[
  datamodel=acmdatamodel,
  style=acmnumeric,
  ]{biblatex}

\usepackage{multirow}
\usepackage{geometry}
\usepackage{graphicx}
\usepackage{subcaption}
\usepackage{float}
\usepackage{amsmath}
\begin{document}

%%
%% The "title" command has an optional parameter,
%% allowing the author to define a "short title" to be used in page headers.
\title{Saliency Driven Imagery Preprocessing for Efficient Compression - Industrial Paper}

%%
%% The "author" command and its associated commands are used to define
%% the authors and their affiliations.
%% Of note is the shared affiliation of the first two authors, and the
%% "authornote" and "authornotemark" commands
%% used to denote shared contribution to the research.
\author{Justin Downes}
%\authornote{Both authors contributed equally to this research.}
\email{jusdow@amazon.com}
\orcid{0000-0003-3257-3359}
\affiliation{%
  \institution{Amazon Web Services}
  \country{USA}
}

\author{Sam Saltwick}
%\authornote{Both authors contributed equally to this research.}
\email{saltwick@amazon.com}
\affiliation{%
  \institution{Amazon Web Services}
  \country{USA}
}
\author{Anthony Chen}
%\authornote{Both authors contributed equally to this research.}
\email{anthche@amazon.com}
\affiliation{%
  \institution{Amazon Web Services}
  \country{USA}
}

%%
%% By default, the full list of authors will be used in the page
%% headers. Often, this list is too long, and will overlap
%% other information printed in the page headers. This command allows
%% the author to define a more concise list
%% of authors' names for this purpose.
\renewcommand{\shortauthors}{Downes et al.}

%%
%% The abstract is a short summary of the work to be presented in the
%% article.
\begin{abstract}
  The compression of satellite imagery remains an important research area as hundreds of terabytes of images are collected every day, which drives up storage and bandwidth costs. Although progress has been made in increasing the resolution of these satellite images, many downstream tasks are only interested in small regions of any given image. These areas of interest vary by task but, once known, can be used to optimize how information within the image is encoded. Whereas standard image encoding methods, even those optimized for remote sensing, work on the whole image equally, there are emerging methods that can be guided by saliency maps to focus on important areas. In this work we show how imagery preprocessing techniques driven by saliency maps can be used with traditional lossy compression coding standards to create variable rate image compression within a single large satellite image. Specifically, we use variable sized smoothing kernels that map to different quantized saliency levels to process imagery pixels in order to optimize downstream compression and encoding schemes.
\end{abstract}

%%
%% The code below is generated by the tool at http://dl.acm.org/ccs.cfm.
%% Please copy and paste the code instead of the example below.
%%


%%
%% Keywords. The author(s) should pick words that accurately describe
%% the work being presented. Separate the keywords with commas.

\received{09 June 2023}
\received[revised]{10 September 2023}
\received[accepted]{28 September 2023 }

%%
%% This command processes the author and affiliation and title
%% information and builds the first part of the formatted document.
\maketitle

\section{Introduction}

Satellite imagery has a long history of novel compression and coding schemes being developed specifically to help optimize transmission from satellite to ground, as well as to minimize storage and processing requirements \cite{compression}. As the analysis of imagery becomes more widespread for machines and humans, the ability to trade fidelity for efficiency through lossy compression becomes more palatable. 

This work demonstrates an approach to preprocessing images that aids arbitrary downstream compression techniques by variably optimizing information encoding for select areas as defined by a given saliency mask. This varies from other techniques as we do not provide an end-to-end solution of variable rate compression but allow for additional preprocessing and a separate compression method. These downstream compression methods may be off-the-shelf image compression standards \cite{1037564}, machine learning based techniques \cite{tfc_github}, or those specifically designed for use with remote sensing \cite{7153923}. Our technique uses smoothing kernels of different sizes that map to quantized saliency levels. These kernels of varying sizes adjust pixel values relative to their neighbors to remove high frequency information which aids in the eventual compression of that pixel. 

In this study we do not focus on generating these saliency maps, as there are a myriad of reasons and methods that users would have interest in specific areas or objects. We instead craft our experiments with the different shapes and sizes these types of saliency maps could reasonably take on. For instance, masking out clouds is a common task \cite{LI2019197}, and would generate maps that could be fairly represented by set of procedurally generated examples. Alternatively, saliency maps driven by interest in certain types of objects \cite{5510681, 6727739} would be reflected by a different set of generated masks. As a first step to this work, we have captured a variety of different methods for saliency generation in the real world and developed abstractions of each method to enable our preprocessing technique to be evaluated on a broad range of input maps. These synthetic saliency maps allow for a variety of underlying image region diversities to have different levels of attention paid to them. 

We utilize standard image reconstruction metrics that compare the decompressed or decoded image to the original image in order to compare storage savings across datasets and saliency mask schemes. We also add evaluations to show the trade-off of information between different regions of importance as well as the impact to the overall image's reconstruction when these masked regions are present. We compare images with variable rates of smoothing to what would be expected of a similar consistent rate of smoothing for a given image to reinforce that information is being sacrificed in low importance areas for that in high.

\section{Background}

Our method relies on smoothing kernels to adjust pixel values in such a way that optimizes downstream compression and encoding. Smoothing kernels, and specifically Gaussian blurring kernels, have had widespread use in a variety of applications in computer vision but have not been widely applied to remote sensing imagery. There has also been focused efforts on maximizing encoding efficiency for remote sensing imagery itself, including the use of saliency masking for targeted compression.

\textbf{Image Encoding.} Specific considerations are necessary for the creation and processing of satellite imagery, including spatial resolution, spectral information, and the trade-off between image-quality and compression ratio \cite{compression}. Traditional and recent methods fall under lossless or lossy categories where the latter is useful in use-cases where a certain degree of distortion is acceptable to the analytic task. 
%For sensitive tasks that necessitate additional data extractions and insights to be processed by computer systems, in on lossless algorithms.

One of these algorithms is JPEG2000 which improved upon its predecessor JPEG with higher compression ratios while achieving lossless image quality \cite{1037564}. The lossless quality from the discrete wavelet-based transformation (DWT) used in JPEG2000 allowed satellite images to retain perceptually significant details allowing for better visual analysis, interpretation, and extraction of relevant information. Along with reduced space for transmission, JPEG2000 also allows satellite imagery to be progressively transmitted over networks with limited bandwidth allowing for low-resolution previews to be usable and accessible in remote areas or limited network conditions. Satellite sensors also often capture multispectral or hyperspectral data which JPEG2000 supports for encoding and compression of data-rich images. Additionally, JPEG2000 allows the inclusion of metadata within the compressed files which is crucial for geospatial analysis to store information such as geolocation data, sensor parameters, and acquisition details. 

The BPG standard \cite{bellard2016bpg} allocates bits based on the homogeneity of the region. Complex regions receive more bits as they are assumed to be more noticeable to encoding artifacts. This rate optimization for more complex regions aligns with our method of preprocessing, but in our method the salient regions are determined externally and not beforehand through hand-crafted metrics.

\textbf{End-to-End ML Methods.} Recently, several ML architectures with built in saliency driven compression have been developed. Encoder-decoder networks have been heavily utilized for compression models and have recently been extended to allow for the integration of task specific quality maps \cite{song2021variable}. These types of architectures typically utilize convolutional layers to extract image features in an encoder model and quality maps to drive variable compression in a spatially-adaptive transformation module. This module compresses the feature maps based on spatial characteristics, effectively allocating more bits to regions with higher visual importance. A decoder module can then reconstruct the compressed image using the compressed features and inverse transforms. 

Another approach to saliency-driven machine learning compression is using a hierarchical auto-regressive model \cite{9423363}. This type of model combines an encoder-outputted importance mask with a generated saliency mask from a pretrained object saliency model. This masking informs the model's ability to compress to a desired target bit-rate value. During the computation of loss, a higher weight is given to the reconstruction of salient regions. 

\textbf{Satellite Imagery Processing.}  In satellite imagery, data quality is important for accurate analysis and interpretation of spatial information. It is also important to optimize processing and compute resource allocations given the large size of these images. Tasks can be optimized by constraining the data based on the information need for that task, for instance masking out clouds \cite{GuEtAl}. Cloud masking focuses on accurately identifying and masking cloud-covered areas, which can hinder various applications, including land cover mapping and change detection. Similar imagery optimizations revolve around removing shadows \cite{rs13040699}. Shadows obscure important details and can greatly affect the accuracy in land cover classification and change detection. In this case the masking is used to produce visually plausible image values by utilizing neighboring information and texture analysis. Many other approaches remove other unwanted artifacts or features through techniques such as water body masking, land cover masking, and urban area masking. All of these masking techniques play the same important role in minimizing certain areas and focusing more on regions of interest. 

Similar to masking, our method can identify and prioritize important regions or features in an area by incorporating saliency into traditional compression. Our approach takes into account the human visual system’s tendency to focus on and perceive details in salient regions, ensuring that critical information is preserved during encoding and possible follow on compression. 

Ensemble approaches to merge different sources of saliency are also used to optimize information encoding \cite{s23020730}. These different sources can allow for flexible methods to derive saliency maps, encompassing techniques from upstream ML models to hand crafted heuristics. Saliency derived from object detection models is one of the proxies we utilized in generating synthetic maps as outlined in \ref{section:mask-generation}. Saliency maps themselves are also used to drive object detection \cite{rs12091435} in certain applications.

\textbf{Scale Space.} The scale space framework is utilized in image processing to understand images at different scales \cite{Lindeberg1994}. This framework uses Gaussian filters to generate images at multiple scales by adjusting the size of the kernel \cite{WITKIN1987329}. This method has been used in compression before \cite{999941}, but this method relied on a chosen scale space representation with fixed kernel size and a novel reconstruction technique. Whereas our method applies variable size kernels within a single image which are determined by a saliency map.

\section{Data}
\begin{figure}[htbp]
    \centering
    \includegraphics[width=\linewidth]{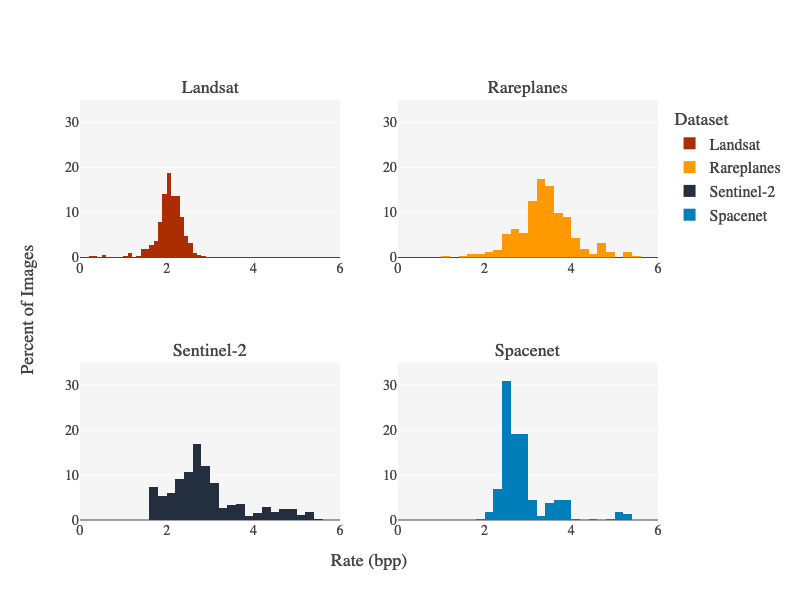}
    \caption{Rate (bpp) distribution across selected datasets.}
    \label{fig:dataset-rates}
\end{figure}

The imagery used in these experiments is curated from a variety of remote sensing datasets. Remote sensing imagery differs from typical images in that they are generally magnitudes larger in dimensions, can have higher bit depth representation for each color, and have varying numbers of color bands. In this effort all images were converted to 3 band, RGB images with 8 bits per color. The distribution of data rate (bpp) for each dataset can be seen in Figure \ref{fig:dataset-rates}. A summary of the datasets used along with the scale of each dataset is provided in Table \ref{tab:dataset-summary}.

\begin{table}[htbp]
    \centering
    \small
    \resizebox{\columnwidth}{!}{\begin{tabular}{|c|c|c|c|c|}
        \hline
        \textbf{Dataset} & \textbf{Subset} & \textbf{Number of Images} & \textbf{Average Width} & \textbf{Average Height} \\
        \hline
        \multirow{2}{*}{Rareplanes} & Real Train + Real Test & 253 & 10,860 & 7,479 \\
        & & & & \\
        \hline
        \multirow{2}{*}{USGS Landsat} & Collection 2, Level-1 ETM & 320 & 8,107 & 7,026 \\
        & & & & \\
        \hline
        \multirow{2}{*}{Sentinel-2} & See Table \ref{tab:sentinel-details} & 339 & 10,980 & 10,980 \\
        & & & & \\
        \hline
        \multirow{2}{*}{Spacenet} & AOIs: 1, 2, 3 & 361 & 8,166 & 8,508 \\
        & & & & \\
        \hline
    \end{tabular}}
    \caption{Dataset Summary with average dimensions of images in pixels. Datasets where filtered to be roughly the same count of images.}
    \label{tab:dataset-summary}
\end{table}

\textbf{RarePlanes} \cite{Shermeyer2021RarePlanesSD} is comprised of a real and synthetic portion, of which only the real images were evaluated in this study. The real portion is made up of 253 satellite scenes (Maxar WorldView-3 \cite{cantrell_system_2021}) covering 112 locations. 

\textbf{SpaceNet} \cite{aws-spacenet} is a repository of freely available imagery with co-registered map features. SpaceNet hosts several different datasets around imagery of 11 global Areas of Interest (AOIs). For this study, a subset of SpaceNet hosted imagery was collected from AOIs 1 (Rio de Janeiro), 2 (Las Vegas), and 3 (Paris). The imagery from these 3 AOIs combines to 361 total images. 

\textbf{Sentinel-2} data is a high resolution land monitoring source of imagery that is refreshed regularly \cite{aws-sentinel2}. Sentinel 2 imagery provides 12 wavelength bands at differing spatial resolutions, subdivided by a UTM tiling grid. Our experiments utilized a sample of the Sentinel-2 L1C dataset for imagery collected in 2023 before May 8th, 2023. Full details on which grids were used can be found in Table \ref{tab:sentinel-details}.

\begin{table}[htbp]
    \centering
    \renewcommand{\arraystretch}{1.0}
    \small
    \begin{tabular}{|l|l|l|l|}
        \hline
        \textbf{UTM Grid Designator} & \textbf{Latitude Band} & \textbf{Square Code} & \textbf{Count} \\ \hline
        17 & S & LA & 28 \\ \hline
        17 & S & MA & 28 \\ \hline
        17 & S & MB & 28 \\ \hline
        17 & S & MD & 28 \\ \hline
        17 & S & NC & 44 \\ \hline
        17 & S & ND & 32 \\ \hline
        18 & S & TH & 51 \\ \hline
        18 & S & TJ & 42 \\ \hline
        18 & S & UH & 28 \\ \hline
        18 & S & UJ & 30 \\ \hline
        \multicolumn{3}{|l|}{\textbf{Total}} & \textbf{339} \\ \hline
    \end{tabular}
    \caption{Sentinel-2 selected areas.}
    \label{tab:sentinel-details}
\end{table}

\textbf{USGS Landsat} is a joint NASA/USGS program that provides the longest continuous space-based record of Earth’s land in existence\cite{aws-landsat}. Landsat is comprised of 8 Earth-observing satellites. The dataset curated for this effort was selected from Landset Collection 2, Level 1 data that was collected by the Enhanced Thematic Mapper (ETM) in 2023.

%We use the imagery from USGS Landsat available through the AWS Registry of Open Data \cite{aws-open-registry}.

%do we want to do xview \cite{1802.07856}

%do we want to do hrsid \cite{hrsid}
\section{Methodology}

Our method utilizes saliency masks to guide spatially-variable image smoothing to control the level of information loss when storing imagery. The saliency masks determine the width of the smoothing function as applied on a per-pixel level, allowing for complete control over the amount of information loss across different regions of an image. 

\subsection{Smoothing Kernel}
\begin{figure}[htbp]
    \centering
    
    \begin{subfigure}{0.40\columnwidth}
        \centering
        \includegraphics[width=\textwidth]{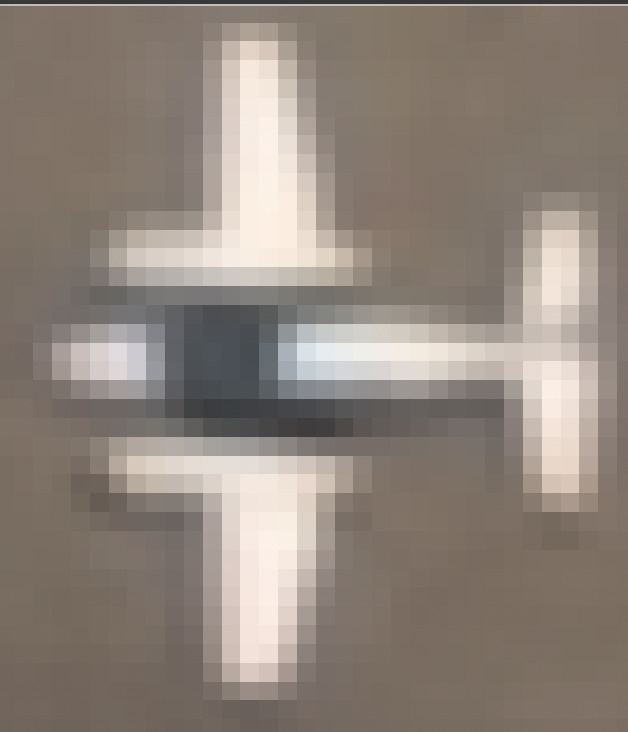}
        \caption{$k=3$}
        
    \end{subfigure}
    \hfill
    \begin{subfigure}{0.40\columnwidth}
        \centering
        \includegraphics[width=\textwidth]{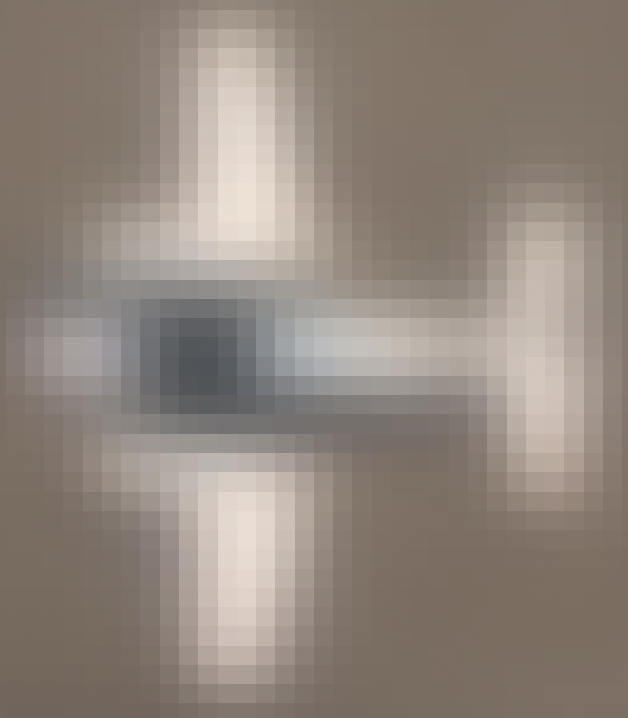}
        \caption{$k=9$}
      
    \end{subfigure}
    \hfill
    \begin{subfigure}{0.40\columnwidth}
        \centering
        \includegraphics[width=\textwidth]{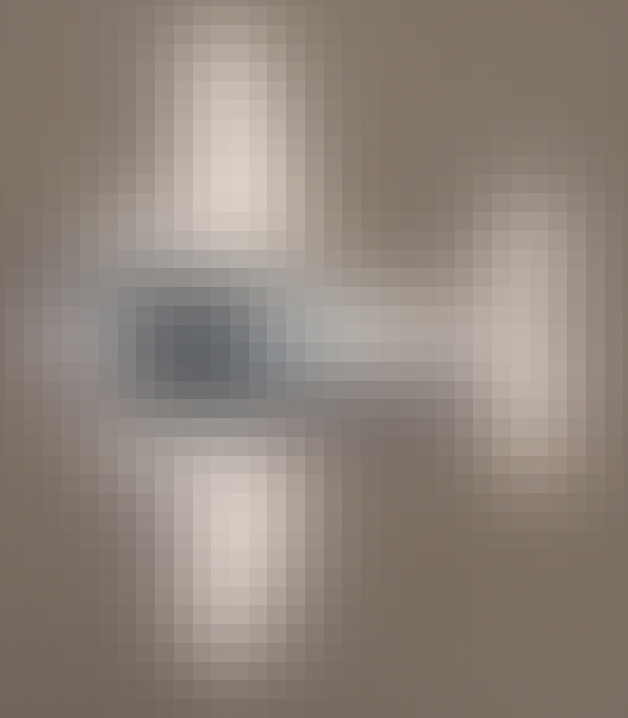}
        \caption{$k=13$}
        
    \end{subfigure}
    \hfill
    \begin{subfigure}{0.40\columnwidth}
        \centering
        \includegraphics[width=\textwidth]{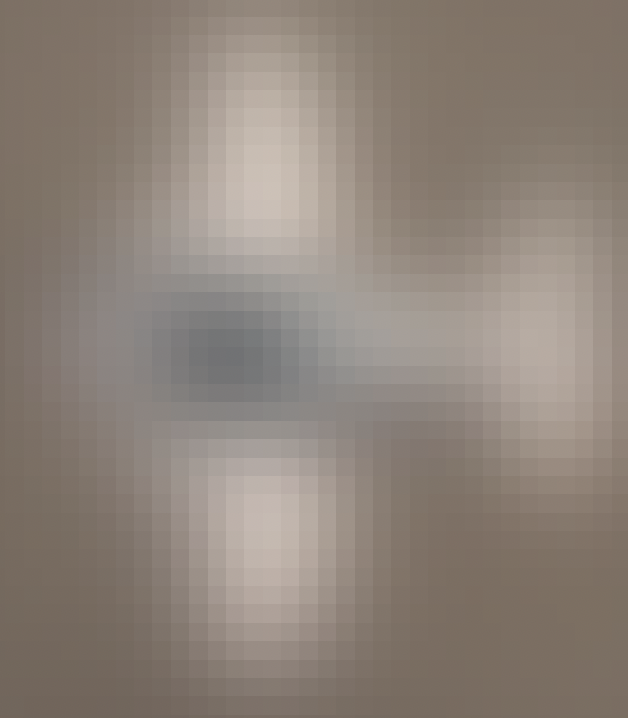}
        \caption{$k=17$}
       
    \end{subfigure}
    
    \caption{Zoomed in crops of various smoothing kernels applied to the whole image.}
    \label{fig:blur-samples}
\end{figure}

Similar to the scale space framework \cite{Lindeberg1994} we utilize a Gaussian kernel with variable sizes to map to different saliency levels. This kernel size is determined and applied per pixel as they directly relate to the saliency map. Shown below in equation \ref{eq:gauss} is the Gaussian smoothing kernel used and in equation \ref{equation:kernel-selection} we show how the kernel sizes are mapped to the Gaussian kernel width ($K(s)$) for a given saliency level $s$. Examples of smoothing kernels applied uniformly to an image are shown in Figure \ref{fig:blur-samples}. 
\begin{equation}
\label{eq:gauss}
    G(x,y) =\frac{1}{2\pi\sigma^2} e^{-\frac{x^2+y^2}{2\sigma^2}}
\end{equation}
\def\lc{\left\lfloor}   
\def\rc{\right\rfloor}
% \begin{equation}
%     x=y\leq \pm \lc\frac{k}{2}\rc,
%     k \in \{3,5,7,9,11,13,15,17\}
% \end{equation}

\begin{equation}
%\begin{align*}
\begin{aligned}
    x=y \leq K(s) \\
    K(s) = 2 \times \lc (1-s) \times \lvert k \rvert \rc + 1, \\
    k \in \{3,5,7,9,11,13,15,17\},\\
    0 \leq s \leq 1
\end{aligned}
\label{equation:kernel-selection}
%\end{align*}
\end{equation}

\subsection{Mask Generation}
\label{section:mask-generation}
\begin{figure}[hbp]
    \centering
    
    %\begin{subfigure}{0.25\textwidth}
    \begin{subfigure}{0.30\columnwidth}
        \centering

        \fbox{\includegraphics[width=\textwidth]{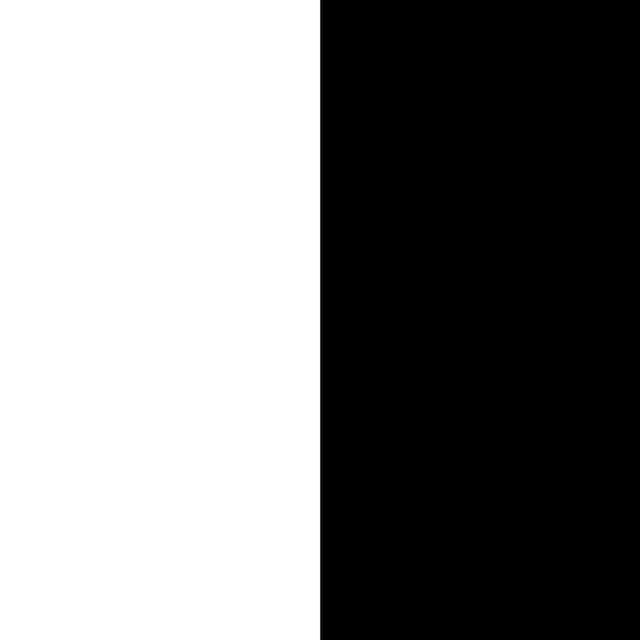}}
        %\fbox{\includegraphics[width=\linewidth]{half_mask.png}}
        \caption{Binary Half \\Mask }
        \label{fig:half-mask}
    \end{subfigure}
    \hfill
    %\begin{subfigure}{0.25\textwidth}
    \begin{subfigure}{0.30\columnwidth}
        \centering

        \fbox{\includegraphics[width=\textwidth]{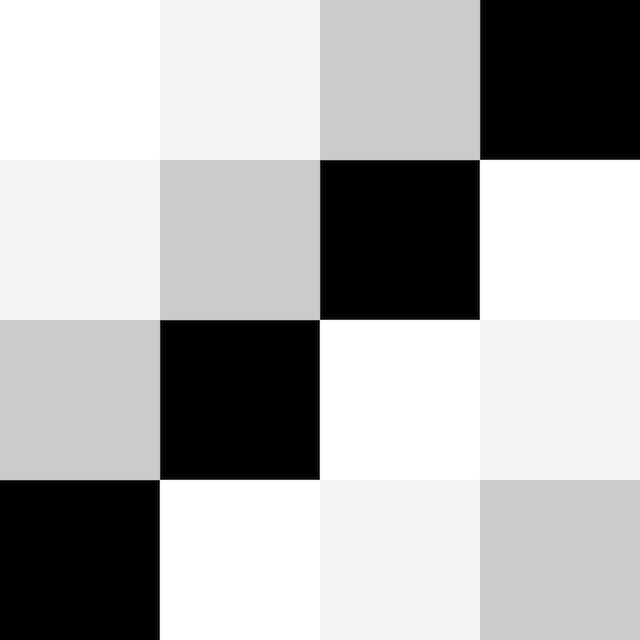}}
        %\fbox{\includegraphics[width=\linewidth]{checker_mask.png}}
        \caption{4x4 Grid \\Mask }
        \label{fig:grid-mask}
    \end{subfigure}
    \hfill
    %\begin{subfigure}{0.25\textwidth}
    \begin{subfigure}{0.30\columnwidth}
        \centering

        \fbox{\includegraphics[width=\textwidth]{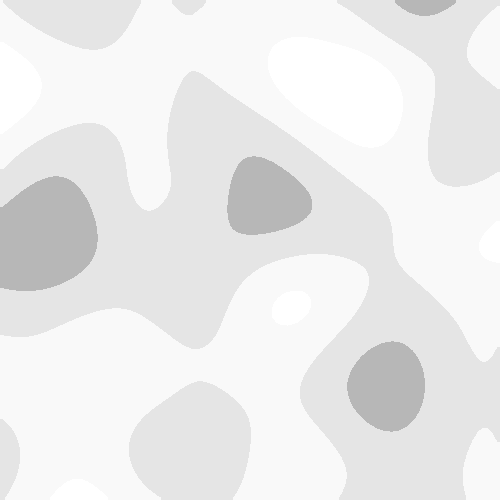}}
        %\fbox{\includegraphics[width=\linewidth]{perlin_mask.png}}
        \caption{Perlin Noise Mask}
        \label{fig:perlin-mask}
    \end{subfigure}
    
    \caption{Generated Saliency Masks}
    \label{fig:mask-fig}
\end{figure}

For these experiments, saliency masks were artificially generated and used to control the spatially variable smoothing kernels. Each saliency mask contains varying normalized saliency levels ranging from 0 to 1. At a saliency level of 0, which is represented by full black in the masks in Figure \ref{fig:mask-fig}, the maximum smoothing kernel is used. At a saliency level of 1, represented by full white in Figure \ref{fig:mask-fig}, no smoothing is applied which preserves the original information.

\begin{figure}[htbp]
    \centering
    
    \begin{subfigure}{0.48\columnwidth}
        \centering
        \includegraphics[width=\textwidth]{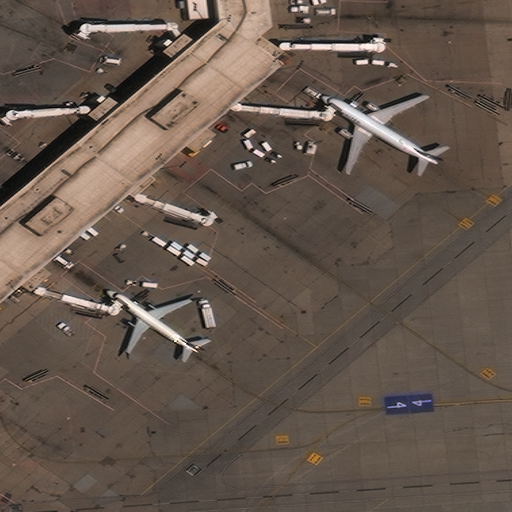}
        \caption{Original Image}
        
    \end{subfigure}
    \hfill
    \begin{subfigure}{0.48\columnwidth}
        \centering
        \includegraphics[width=\textwidth]{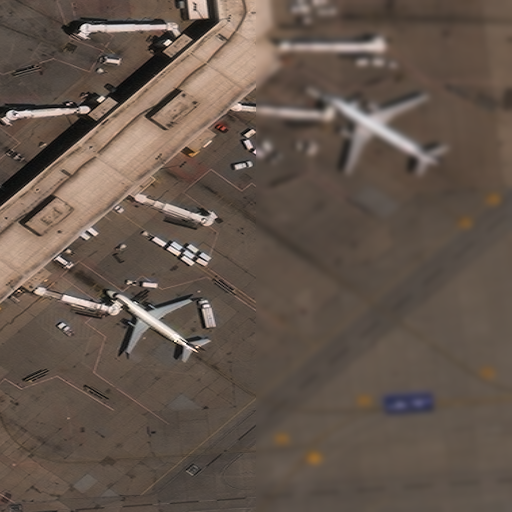}
        \caption{Half Mask}
      
    \end{subfigure}
    \hfill
    \begin{subfigure}{0.48\columnwidth}
        \centering
        \includegraphics[width=\textwidth]{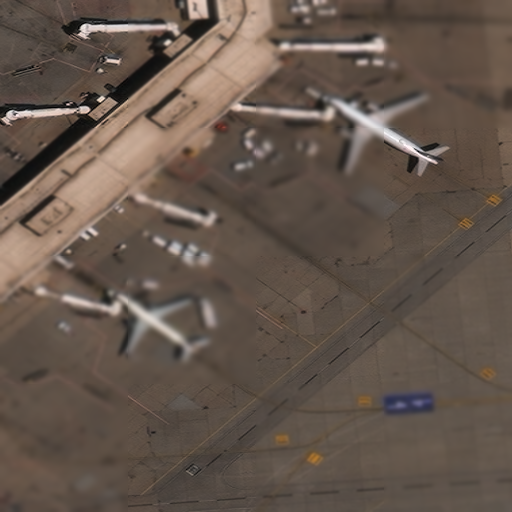}
        \caption{Grid Mask}
        
    \end{subfigure}
    \hfill
    \begin{subfigure}{0.48\columnwidth}
        \centering
        \includegraphics[width=\textwidth]{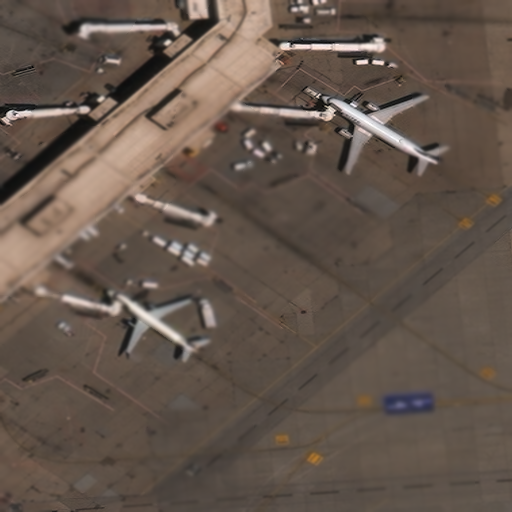}
        \caption{Perlin Mask}
       
    \end{subfigure}
    
    \caption{Examples of various saliency-masks applied with spatial-blurring.}
    \label{fig:spatial-blur-samples}
\end{figure}

Three different masks were generated to simulate different use cases. First, a binary mask with maximum saliency (a saliency level of 1) on the left half and minimum saliency (a saliency level of 0) on the right half is used to preserve full information in one half of the image while maximally reducing information in the rest of the image (Figure \ref{fig:half-mask}). The second mask consists of a 4x4 grid consisting of saliency levels [0, 0.33, 0.66, 1] in a rotating pattern (Figure \ref{fig:grid-mask}). The final mask is a grid of evenly sampled perlin noise generated with 3 octaves. The noise is scaled to the range 0-255 to represent an 8-bit image and then reduced to 2-bits by truncating the 6 least significant bits. This results in 4 evenly spaced values, which are then directly mapped to the normalized saliency range of 0 to 1. The perlin noise mask used in these experiments consists of saliency levels [0.25, 0.50, 0.75, 1.0] (Figure \ref{fig:perlin-mask}). Examples of the masks applied to an image can be seen in Figure \ref{fig:spatial-blur-samples} 

Each mask contains a different distribution of saliency levels across its pixels. Saliency levels map to smoothing kernel widths based on Equation \ref{equation:kernel-selection}. The distribution of kernel widths mapped to by each mask can be found in Table \ref{tab:mask-distribution}. 

\begin{table}[htbp]
    \centering
    \renewcommand{\arraystretch}{1.0}
    \small
    \begin{tabular}{|l|l|l|}
        \hline
        \textbf{Mask} & \textbf{Kernel Width} & \textbf{Pixel Distribution} \\ \hline
        Grid & [1, 5, 11, 17] & [0.25, 0.25, 0.25, 0.25] \\ \hline
        Half & [1, 17] & [0.50, 0.50] \\ \hline
        Perlin & [1, 5, 9, 13] & [0.06, 0.44, 0.42, 0.08] \\ \hline
    \end{tabular}
    \caption{Kernel Width Distribution across Mask Types}
    \label{tab:mask-distribution}
\end{table}

%\begin{figure*}[h!]
%\begin{center}
%{\includegraphics[width=0.95\linewidth]{ObjectFocusMaskGraphic.png}}
%\caption{An object focused saliency mask applied to an image.}
%\label{object-workflow}
%\end{center}
%\end{figure*}

%\begin{figure*}[h!]
%\begin{center}
%{\includegraphics[width=0.95\linewidth]{ObjectFocusMaskGraphic4.png}}
%\caption{An object focused saliency mask applied to an image.}
%\label{object-workflow}
%\end{center}
%\end{figure*}

\subsection{Saliency Driven Smoothing}

In order to control the amount of smoothing based on levels defined by the saliency masks, a subset of available smoothing kernels are selected and applied to the image. First, the mask is scaled to match the exact dimensions of the target image. Next, the normalized saliency levels in the mask ([0-1]) are mapped to a value of k, as described in Equation \ref{equation:kernel-selection}, to create a set of  \emph{N} smoothing kernel sizes. A smoothing function of each kernel size is then applied to the image in full to create \emph{N} versions of the image at varying levels of smoothness. Finally, for each of the \emph{N} kernel widths, a binary mask is created to identify the pixels in the image that should be smoothed with that kernel and all \emph{N} smoothed images are composited according to the set of \emph{N} binary masks. This results in a single image with pixel-level control over which regions are smoothed according to the saliency mask. The computational complexity of this operation is linear with the number of unique kernel widths specified by the saliency mask and bounded by the imposed set of available kernel widths $k$. Figure \ref{fig:spatial-blur-samples} shows examples that have been blurred using each of the three generated saliency masks and our spatial blurring method. 

Since the masks used in saliency-driven smoothing can utilize multiple kernel widths, we calculate an effective kernel width \emph{$K_{e}$} for the entire mask \emph{M} using Equation \ref{eq:effective-kw}. In summary, the effective kernel size is an average of kernel sizes mapped by the mask, weighted by the ratio of the size of the masked areas to the size of the image. We use this effective kernel width to compare saliency-driven smoothed composites with uniformly smoothed images. 

\begin{equation}
\label{eq:effective-kw}
    K_{e} (M) = \sum_{k \in M_{\text{unique}}} w_k \times k,
    w_k = \frac{|M_{M=k}|}{|M|}
\end{equation}

\subsection{Encoding and Compression}
After the composite image has been constructed from the various levels of saliency-driven smoothing, the image can be written to storage in an efficient manner through an image encoding method. As different encoding methods exploit different image features to produce compact representations, smoothing as a pre-processing method may affect them differently. In order to evaluate these effects, we studied smoothing as a pre-processing technique to two different image encoding mechanisms. The software package ImageMagick \cite{imagemagick} was utilized to perform both methods.

We study the effects on JPEG2000, which utilizes a discrete wavelet-based transformation (DWT) and DWT coefficient quantization. The ImageMagick interface for JPEG2000 File Format (JP2) includes a target compression rate argument named \emph{rate}, where a \emph{rate} of 1 is equivalent to JPEG2000 lossless compression. We also study the effects on the Better Portable Graphics (BPG) format, based on the intra-frame encoding of High-Efficiency Video Coding (HEVC) \cite{bellard2016bpg}. The ImageMagick interface for BPG includes a \emph{quality} argument that controls the quantization parameter of the HEVC encoding. The quantization of frequency component coefficients is a function of this quantization parameter \cite{1218200}.

\section{Results}

We evaluate the impact that our saliency-based smoothing method has on both image quality and space savings. We use mean squared error (MSE) to evaluate image quality. The image encoding storage rate as bits per pixel (bpp) and the percent reduction in bpp are used to evaluate the effects on image file size. 

\subsection{Saliency Based Smoothing}

We evaluate our saliency-based smoothing by comparing saliency-based smoothed composite images to both the original images and to uniformly blurred images at varying smoothing kernel widths. Images compared in this section were all stored as PNG files to maintain the same encoding as the original imagery. The effective kernel width, described in Equation \ref{eq:effective-kw}, is used to directly compare the qualities of uniformly smoothed images and spatially smoothed composite images. Since we know the saliency masks degrade quality for less important regions, this comparison shows that storage savings are being achieved solely in those regions and is comparable as to having blurred the entire image.

\begin{figure}[h]
    \centering
    \begin{subfigure}{0.95\columnwidth}
    %\begin{figure}[htbp]
        \centering
        \includegraphics[width=\textwidth]{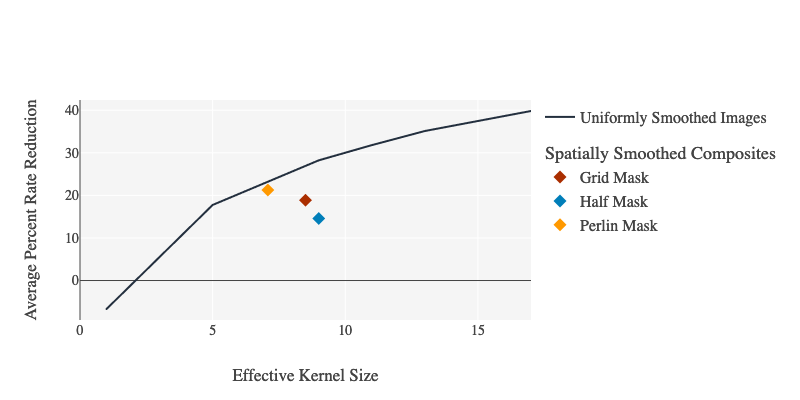}
        \caption{Comparison of Storage Rate Reduction for Uniform and Saliency-Based Smoothing}
        \label{fig:effective-k-rr}
    \end{subfigure}

    \begin{subfigure}{0.95\columnwidth}
        \centering
        \includegraphics[width=\textwidth]{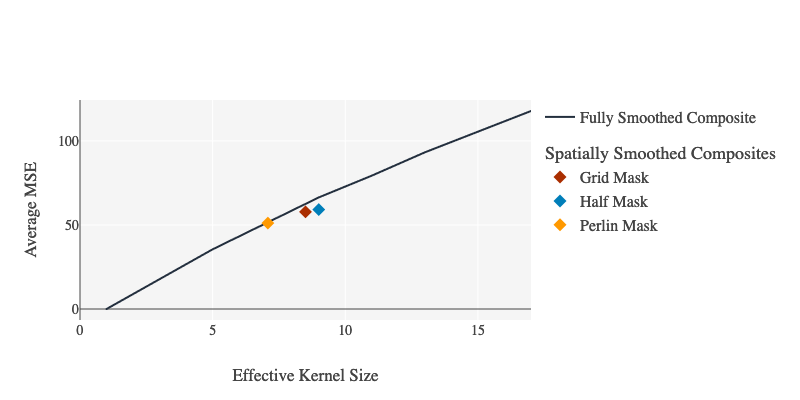}
        \caption{Comparison of Error for Uniform and Saliency-Based Smoothing}
        \label{fig:effective-k-mse}
    \end{subfigure}
    \caption{Comparison of spatially smoothed composites to their uniformly smoothed counterparts. Since spatially smoothed composites target low saliency regions for higher compression, we show effective trade-off of information encoding between low and high imporatance regions }
\end{figure}

Figure \ref{fig:effective-k-rr} depicts the relationship between effective kernel size and average percent storage rate reduction for both uniformly and spatially smoothed images. For uniformly smoothed images, the rate reduction scales in a logarithmic manner with kernel size. Across our data and saliency masks, spatially-smoothed composite images have a lower rate reduction at an equivalent kernel size than uniformly smoothed images. Similarly, Figure \ref{fig:effective-k-mse} illustrates the relationship between effective kernel size and average visual error (MSE). The relationship between kernel size and MSE for uniformly smoothed images is approximately linear. Across our data and saliency masks, we observe that the spatially-smoothed images have a lower MSE than images uniformly smoothed at the same kernel size as the effective kernel size in the spatially-smoothed image. While the rate reduction and error are highly variable depending on the contents of the image, we believe this relationship highlights the trade off between image quality and space savings when spatially-blurring imagery. 

\begin{figure*}[h!]
    \centering
    \includegraphics[width=.95\linewidth]{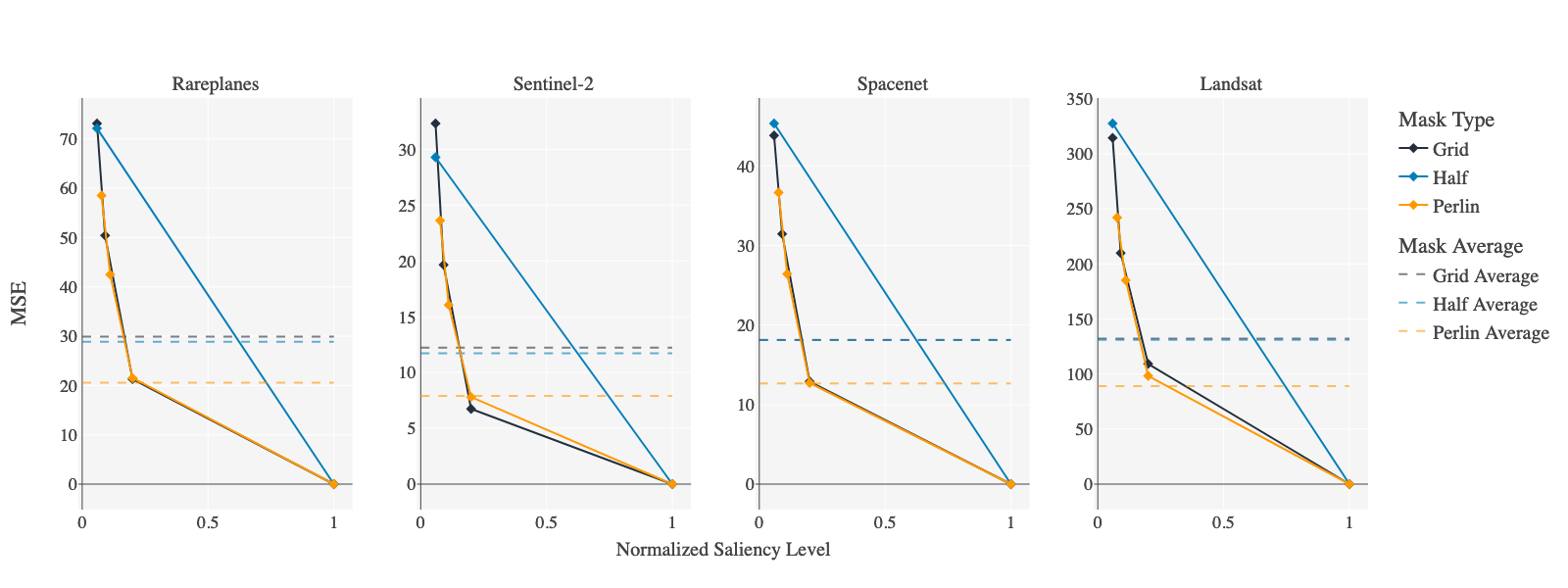}
    \caption{Varying image quality within saliency masks. Saliency level of 1 means the pixel values were not adjusted. We show that higher saliency levels have lower error rates with our method. Note, the different scales of MSE for each dataset.}
    \label{fig:varying-quality}
\end{figure*}

\begin{figure}[H]
    \centering
    \includegraphics[width=\linewidth]{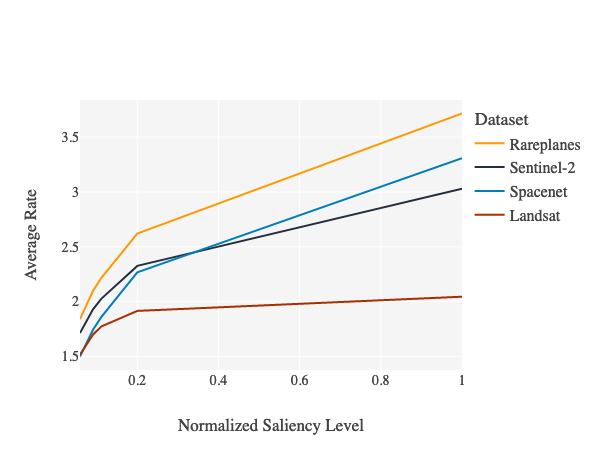}
    \caption{Relationship between normalized saliency level and storage rate reduction.}
    \label{fig:saliency-rate}
\end{figure}

As described in Section \ref{section:mask-generation}, each mask contains varying saliency levels and therefore results in varying error rates across pixels in the image. Figure \ref{fig:varying-quality} shows this relationship across the four datasets evaluated. At every saliency level in each mask, the corresponding region of the spatially-smoothed image can be compared to the original to identify the spatially-aware error. Figure \ref{fig:varying-quality} shows that within the spatially-smoothed imagery, there are regions with a saliency level of 1, corresponding to no loss of information or zero error. At the same time, rate savings are introduced through regions of the image with lower saliency levels, and therefore higher error rates and higher rate reduction. Figure \ref{fig:varying-quality} also shows this relationship for each saliency mask type, where the number of points on each curve represents the number of unique saliency levels in the mask. The relationship between the saliency levels and the average rate across datasets is observed in Figure \ref{fig:saliency-rate}. At extremely low saliency levels, the average rate decreases by up to 47.9\%. At the highest saliency levels, while there is no error, the average rate remains at or above the original data rate. Observed increases in rate are likely due to differences between source and destination encoding as well as the re-encoding of images as PNGs. Across the four datasets used, we observed significant variance in the relationship between saliency level and rate, which we attribute to differences in the image information and it's ability to be reduced by Gaussian blurring.

\begin{figure}[H]
    \centering
    \includegraphics[width=\linewidth]{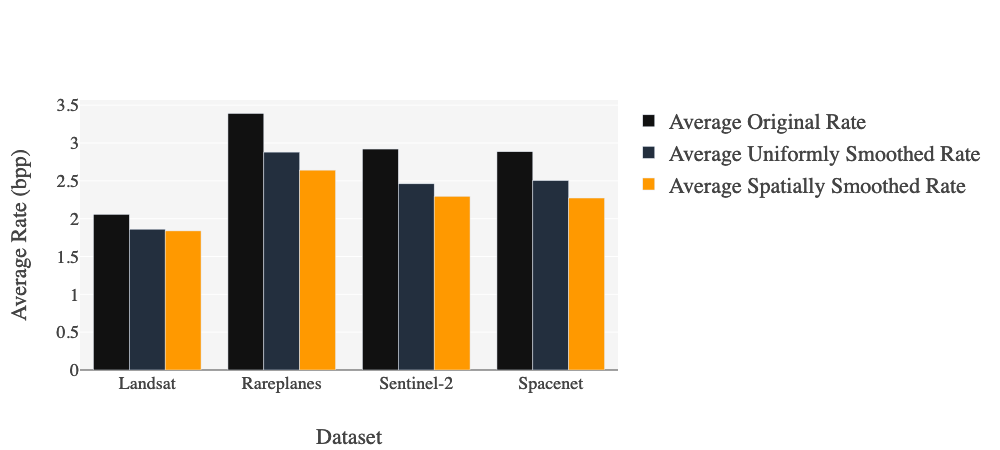}
    \caption{Effect of smoothing on storage rate by dataset}
    \label{fig:rate-dataset}
\end{figure}

\begin{figure}[H]
    \centering
    \includegraphics[width=\linewidth]{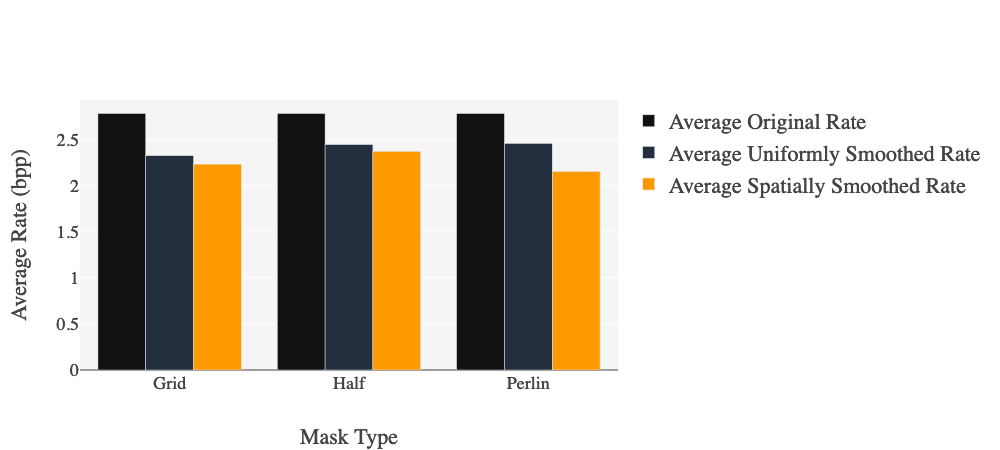}
    \caption{Effect of smoothing on storage rate by mask type}
    \label{fig:rate-mask}
\end{figure}

Figure \ref{fig:rate-dataset} depicts the effects of both uniformly and spatially smoothing imagery across observed datasets, where the average spatially smoothed rate is computed across all 3 mask types. Uniform smoothing results in slightly reduced data rates and the composite smoothing results in even further reductions on average. We believe that the magnitude of rate reduction observed here is highly dependant on image contents and what patterns in the image are being smoothed. Similarly, Figure \ref{fig:rate-mask} shows reductions in data rate from smoothing across the three saliency-based masks used. The reduced rate between the uniformly blurred and spatially-blurred imagery is again likely due to the contents of the image being spatially blurred. However, the perlin mask results in a larger reduction in data rate, due to the distribution of kernel widths used, as described in Figure \ref{tab:mask-distribution}.

\subsection{Impact on Downstream Compression}

\begin{figure*}[htbp]
    \centering
    \includegraphics[width=\linewidth]{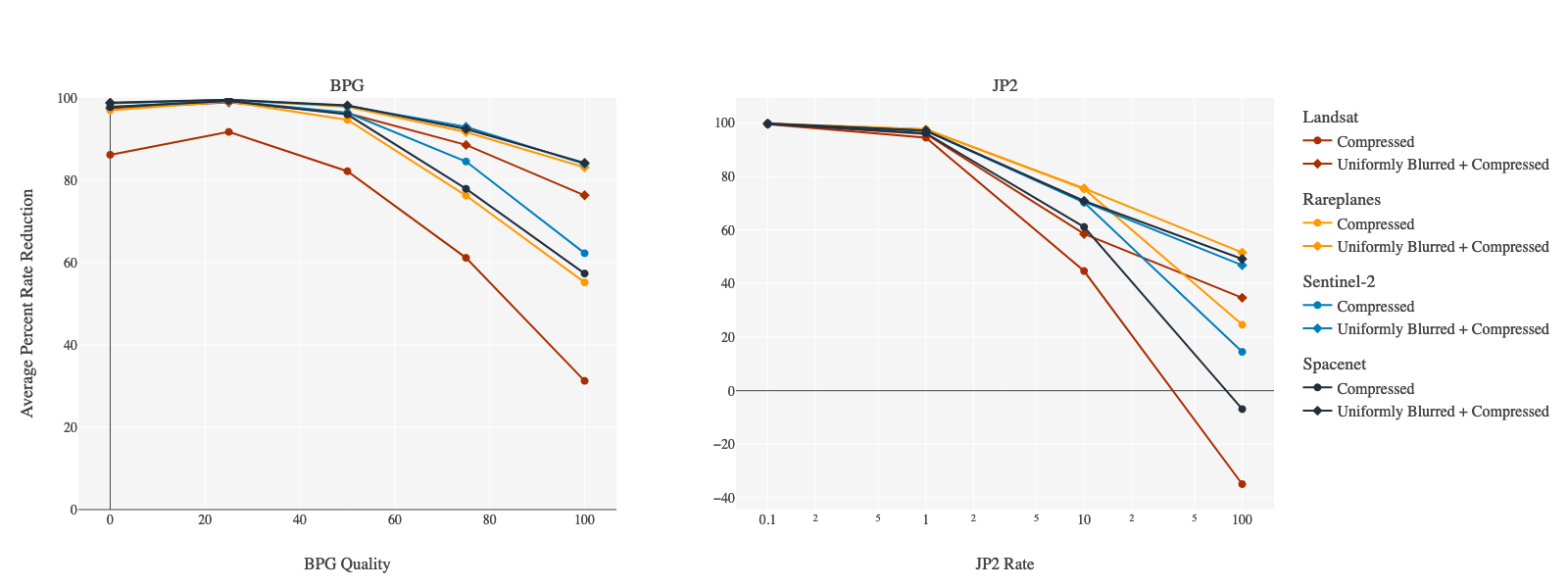}
    \caption{Impact of uniform smoothing on BPG and JP2 encoding by compression parameter for storage rate reduction. Negative rate reductions mean the image storage size increased. Note, JP2 rates were rescaled for consistency with BPG quality values.}
    \label{fig:ub-on-compress}
\end{figure*}

\begin{figure*}[htbp]
    \centering
    \includegraphics[width=\linewidth]{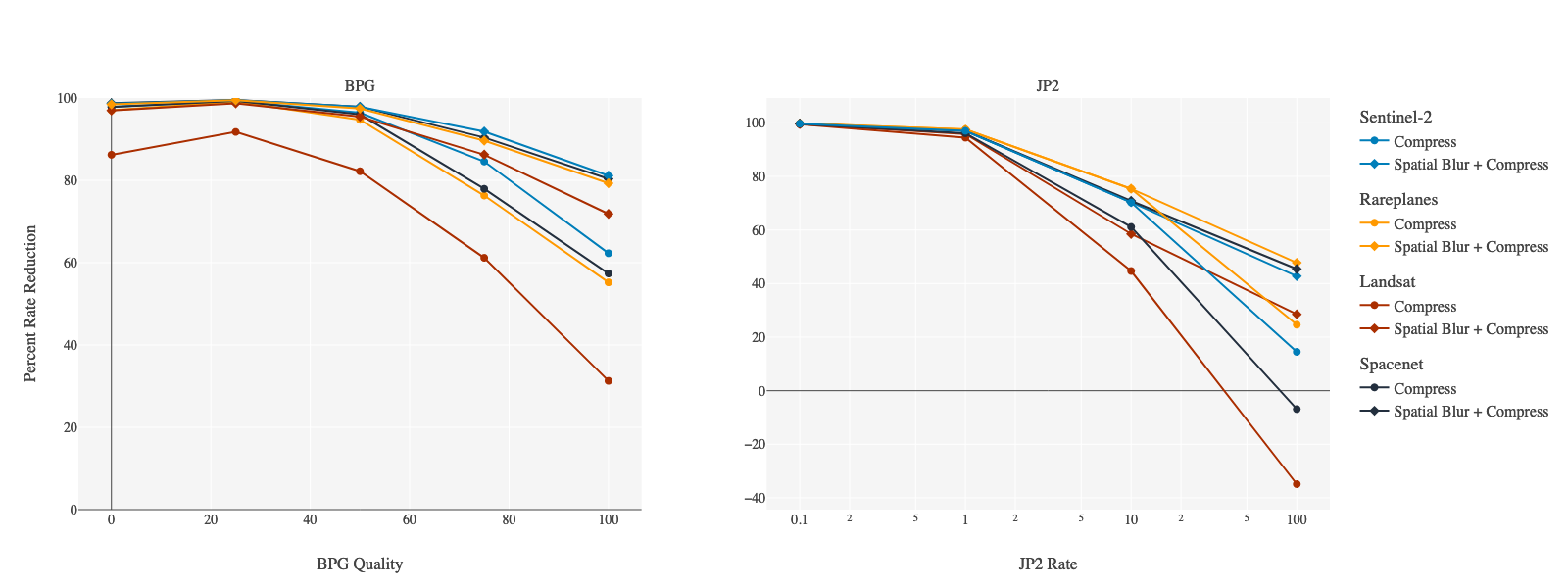}
    \caption{Impact of saliency-driven smoothing on BPG and JP2 encoding by compression parameter for storage rate reduction. Negative rate reductions mean the image storage size increased. Note, JP2 rates were rescaled for consistency with BPG quality values.}
    \label{fig:sdb-on-compress}
\end{figure*}

\begin{figure*}[htbp]
    \centering
    \includegraphics[width=\linewidth]{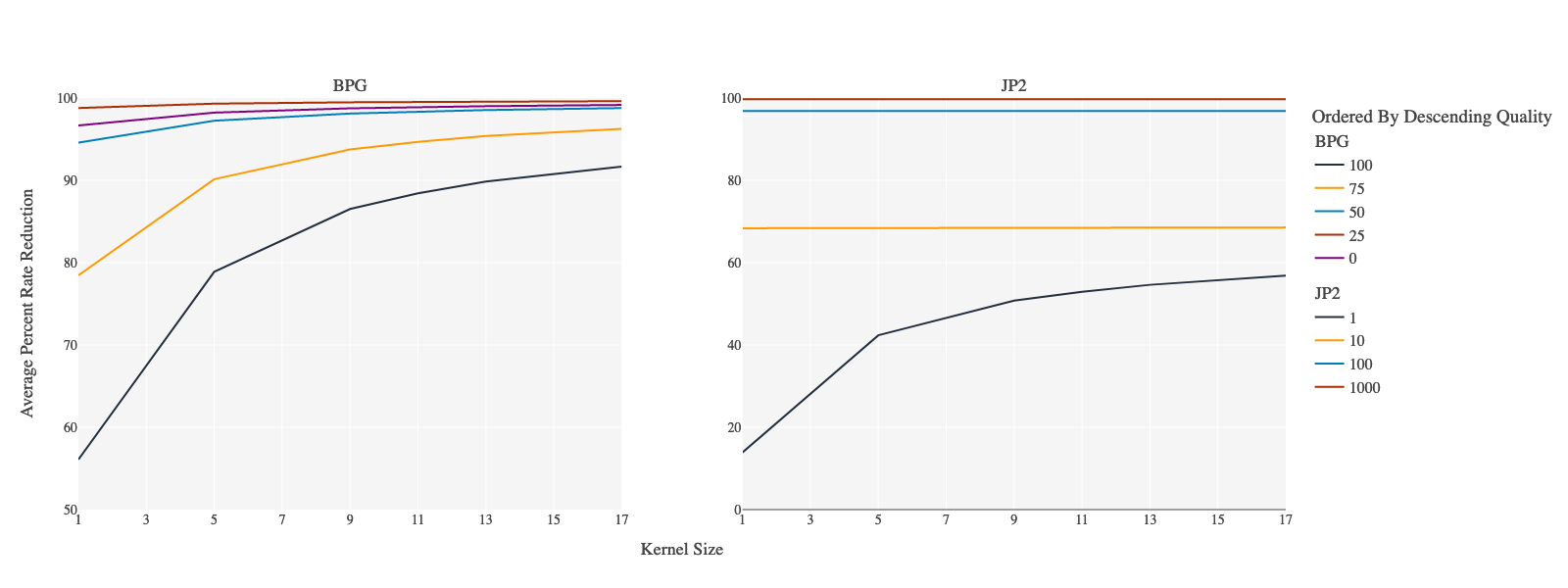}
    \caption{Impact of uniform smoothing kernel size on BPG and JP2 compression for storage rate reduction.}
    \label{fig:k-on-rr}
\end{figure*}

In addition to the impacts of smoothing on same-encoding image storage, we study the impacts of smoothing on further compression of imagery. In particular, we single out JPEG2000 and BPG as compression methods to evaluate against. While the  JPEG2000 rate parameter usually represents an estimated rate reduction, we invert and rescale the rate values, so that a value of 100 represents a JPEG2000 rate of 1, for consistency with the BPG quality parameter. Figure \ref{fig:ub-on-compress} shows the impact of uniform smoothing on both compression methods. Here, we observe that uniform smoothing has a larger impact on compression at higher BPG quality levels and JPEG2000 rate levels. Uniform smoothing results in significant rate reductions at these low amounts of compression, up to a 45\% difference for the Landsat imagery. At higher levels of compression, the effect diminishes and smoothing carries little impact at extremely high levels of compression. We observed that the impacts at varying compression levels are both data and encoding scheme dependent, as the rate difference between smoothed and unsmoothed compressed imagery varied across tested datasets, even at extremely low BPG qualities.

We then studied the impact of our saliency-driven spatial smoothing on further image compression. Results in Figure \ref{fig:sdb-on-compress} are almost identical to the uniform smoothing, following a similar pattern with larger effects at lower magnitudes of file compression. We believe this pattern shows that spatially-compressing imagery does not negatively impact the ability of compression methods to be improved by smoothing and that gaussian smoothing does provide a mechanism to optimize compression encoding methods.

When controlling for uniform smoothing kernel size, we observe similar results. Smoothing kernel size has a significantly larger impact on rate reduction at low levels of compression. At extremely high levels of compression, the effects of uniform smoothing are significantly diminished. Similar to Figure \ref{fig:ub-on-compress}, the impacts at this extreme vary across sets of imagery and compression methods. We observed that at high kernel sizes, rate reduction nearly converges for BPG, but not for JPEG2000, as shown in Figure \ref{fig:k-on-rr}. 

\begin{figure}[H]
    \centering
    
    \begin{subfigure}{0.48\columnwidth}
        \centering
        \includegraphics[height=3cm]{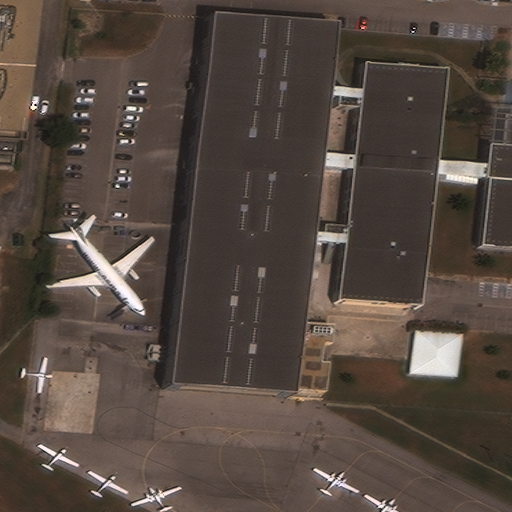}
        \caption{Original Image}
        
    \end{subfigure}
    \hfill
    \begin{subfigure}{0.48\columnwidth}
        \centering
        \includegraphics[height=3cm]{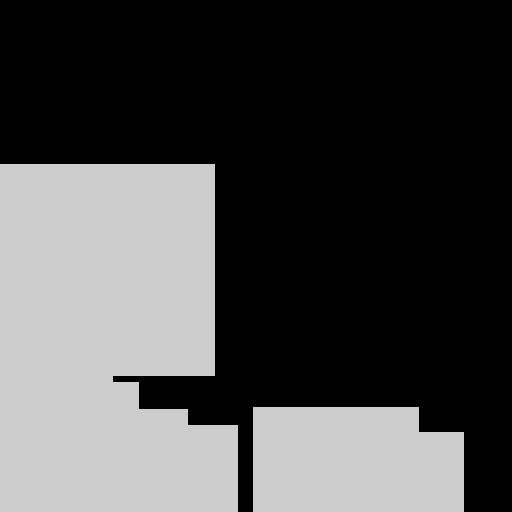}
        \caption{Saliency Mask}
      
    \end{subfigure}
    \hfill
    \begin{subfigure}{0.70\columnwidth}
        \centering
        \includegraphics[height=4cm]{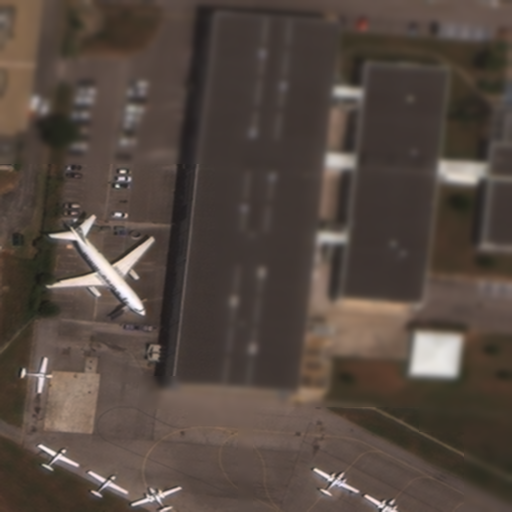}
        \caption{Saliency-Smoothed Composite}
        
    \end{subfigure}
    
    \caption{Example simulated saliency mask created from objects of interest and applied to an image.}
    \label{fig:detection-example}
\end{figure}

We also study the effects of saliency-driven spatial smoothing on a downstream task to evaluate any changes in performance due to variable smoothing. We use the test subset of the RarePlanes imagery along with the pre-trained Faster-RCNN \cite{ren2016faster} object detector provided by the RarePlanes curators. Simulated saliency masks are generated by using the ground truth detection boxes, significantly dilating each box individually, and creating a binary mask with a normalized saliency value of 0.8 within the dilated box and 0 outside of it. An example saliency mask can be seen in Figure \ref{fig:detection-example}. We utilize the Detectron2 library \cite{wu2019detectron2} for model inference and evaluation and study the effects on mean average precision (AP), AP at 50 (AP50), AP at 75 (AP75), AP on small objects (APs), AP on medium objects (APm), and AP on large objects (APl). Object sizes are determined by the Detectron2 COCOEvaluation module.

\begin{figure}[H]
    \centering
    \includegraphics[width=\linewidth]{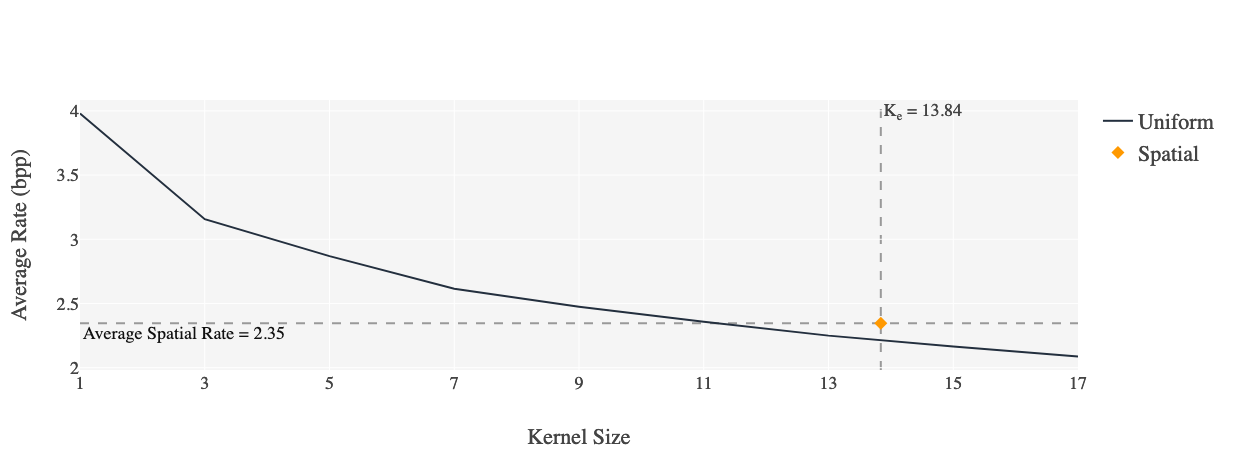}
    \caption{Average bits per pixel (BPP) rate vs. kernel size for uniformly and spatially smoothed RarePlanes-test imagery}
    \label{fig:detection-rate-kernel}
\end{figure}

\subsection{Impact on Downstream Object Detection}

In addition to the simulated saliency masks, the RarePlanes-test imagery is uniformly smoothed at the entire range of kernel sizes. Figure \ref{fig:detection-rate-kernel} shows the relationship between smoothing kernel size and resulting rate averaged across the entire dataset. The singular point in the figure represents the spatially-smoothed imagery with an effective kernel size of $K_{e}=13.84$. The slight increase in bit rate compared to an equivalent kernel size used to uniformly smooth the images highlights the trade-off being made, where parts of the image are stored at a higher bit-rate.

\begin{figure}[H]
    \centering
    \includegraphics[width=\linewidth]{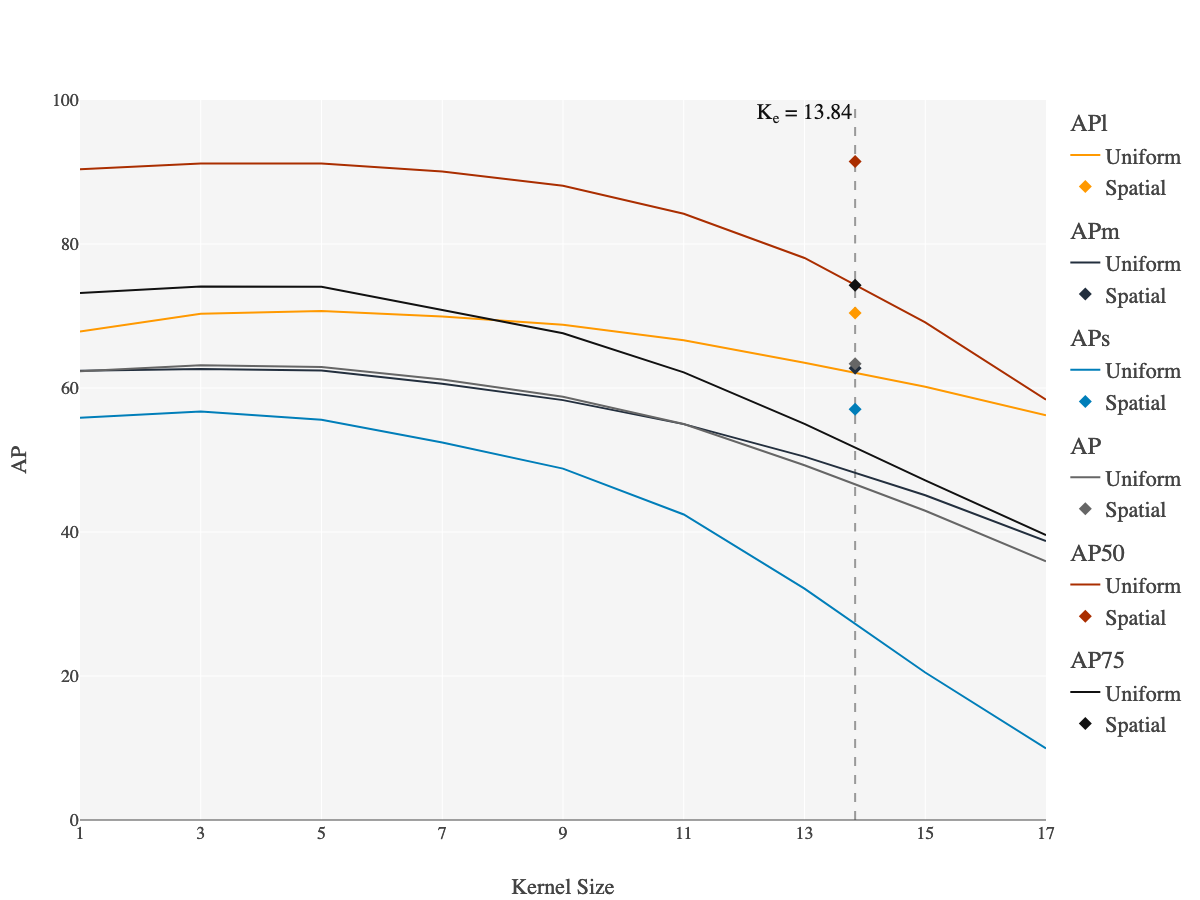}
    \caption{Comparison of detection performance with uniform and spatial smoothing by kernel size.}
    \label{fig:detection-kernel-size}
\end{figure}

We evaluate the object detection model on the uniformly smoothed imagery to establish a baseline. Figure \ref{fig:detection-kernel-size} depicts the observed relationship between kernel size and AP scores, with increased kernel sizes reducing model performance. The only exception to this relationship is at a kernel size $K=3$, where object detection performance slightly increases at a maximum of 3.50\% for AP-large. In comparison, the detection results improved across all metrics for the spatially-smoothed imagery, not just at an equivalent kernel size, but above the peak model performance at $K=3$. 

\begin{figure}[H]
    \centering
    \includegraphics[width=\linewidth]{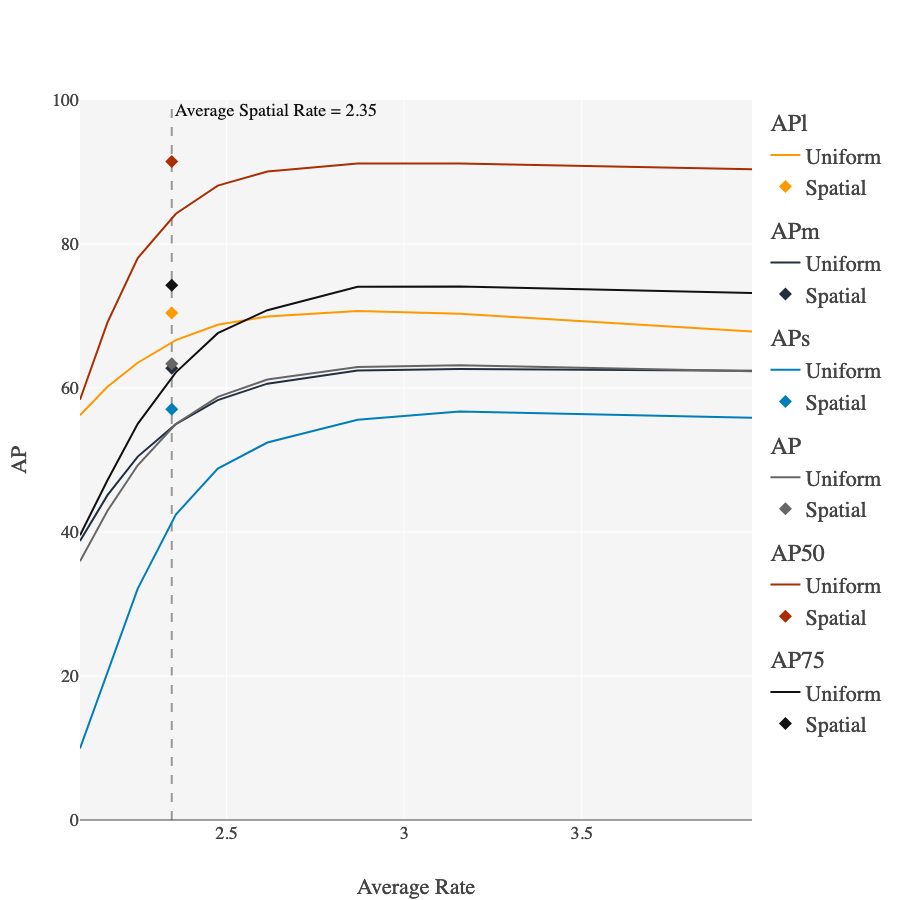}
    \caption{Comparison of detection performance with uniform and spatial smoothing by average rate.}
    \label{fig:detection-rate}
\end{figure}

We also compare detection performance at equivalent bit-rate. For uniformly smoothed imagery, detection performance is greatly impacted by significantly reduced rates, as seen in Figure \ref{fig:detection-rate}. However, detection on the spatially-smoothed imagery is not impacted by the rate reduction, since the objects of interest did not have their rate, and therefore quality, significantly reduced. In fact, since a small amount of smoothing was applied to the regions of interest ($K=3$), there is a slight improvement in performance over a completely unsmoothed baseline ($K=1$ in Figure \ref{fig:detection-kernel-size}). This further reinforces that for targeted tasks, saliency driven preprocessing can more efficiently encode an image while preserving the performance on that downstream task.

\begin{figure}[H]
    \centering
    \includegraphics[width=\linewidth]{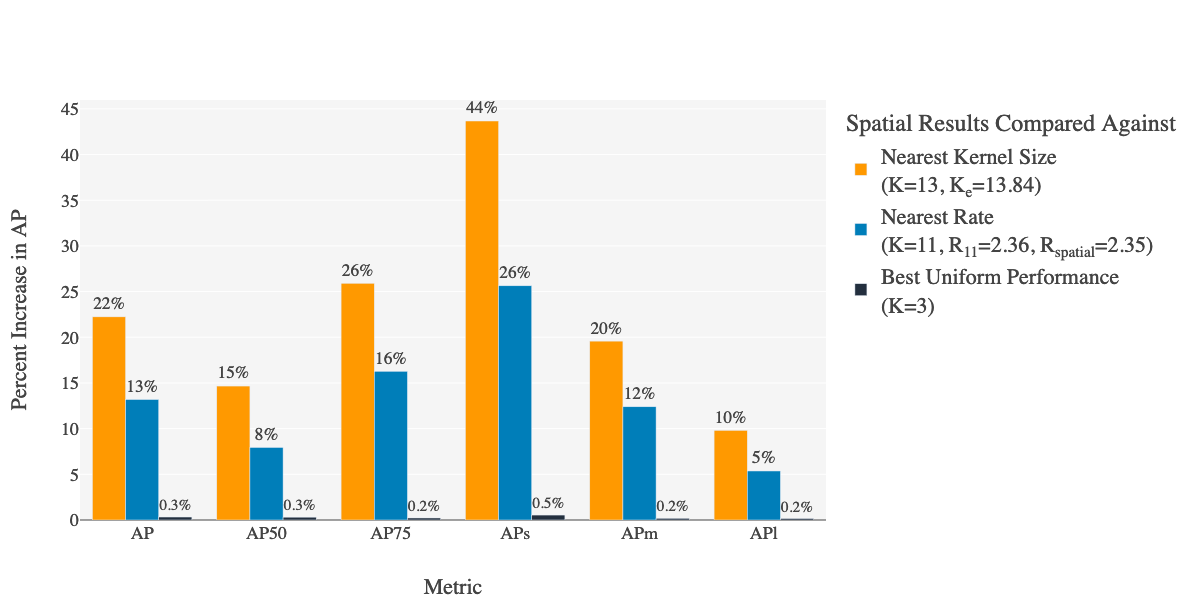}
    \caption{Comparison of detection performance of object focused, spatially smoothed images to their nearest baseline; kernel size, storage rate, and the best uniformly smoothed performances.}
    \label{fig:detection-comparison}
\end{figure}

In Figure \ref{fig:detection-comparison}, we directly compare object detection performance across several of the baselines and find that the spatially-smoothed imagery results in an increase in performance across all baselines. When compared to imagery uniformly smoothed at the nearest kernel size ($K=13$ vs. $K_{e}=13.84$), there are increases in performance upwards of 43.7\%. When compared to imagery stored at the nearest bit-rate, uniformly smoothed with kernel size $K=11$ resulting in a bit-rate of $R_{11}=2.36$ vs $R_{spatial}=2.35$,  detection AP improves by upwards of 25.6\%. Finally, when compared to the best performing baseline ($K=3$), we find there is still a slight improvement in detection performance at an increase of AP-small of 0.5\%. 

Overall, we find that saliency-driven spatial smoothing can significantly improve object detection results at equivalent imagery bit-rates. We also find that a minimal amount of image smoothing may increase object detection performance, resulting in performance gains on both minimally uniformly smoothed and spatially smoothed imagery.

\section{Conclusion}
We introduce a saliency-driven image preprocessing technique for efficient compression of satellite imagery. Saliency masking in conjunction with variable width smoothing kernels construct spatially-blurred composite images, which maintain the quality of salient image regions while introducing additional storage rate savings. Further study is warranted on the impacts of saliency-driven image smoothing relative to the contents of imagery, as we believe this introduced variance across our experiments. These saliency-driven spatially blurred images can then be further compressed through traditional compression and encoding schemes. The combination of saliency-driven blurring and downstream compression results in an optimized method for preserving salient region quality while maximizing file storage savings.

%%
%% The acknowledgments section is defined using the "acks" environment
%% (and NOT an unnumbered section). This ensures the proper
%% identification of the section in the article metadata, and the
%% consistent spelling of the heading.
\begin{acks}
The authors would like to thank project collaborators Joseph Fahimi, Newel Hirst, and Kelvin Yuen.
\end{acks}

%%
%% Print the bibliography
%%
\printbibliography

%%
%% If your work has an appendix, this is the place to put it.
\appendix
\section{Industrial Applications}
Optimizing the utilization of storage and bandwidth is a strong requirement for any application working with remote sensing data at scale. The large image sizes coupled with narrow areas of interest, leaves plenty of room for introducing efficiencies. Some of the use-cases that inspired or could be aided by this work are outlined here.

\textbf{Disconnected Edge.}
Storage of imagery data at the edge is useful in disconnected settings for numerous use cases. For instance, in post-disaster scenarios where traditional infrastructure is disrupted or unavailable, such as the earthquakes in Haiti \cite{sheller}. Although there were many institutions with advanced technologies for satellite imaging of topographical maps of Haiti, very few Haitians were able to access them. There was uneven connectivity and network coverage that made broadband access to these satellite images problematic for decision making, planning, and rebuilding. Even highly developed countries such as Japan during the 2011 earthquake and nuclear meltdown can experience unequal network connectivity between urban and rural areas. By optimizing image encoding to focus on only those portions of the image that are important to the situation more images can be stored locally and I/O usage can be minimized.

Work is also being done to optimize compression specifically at the edge. Unlike traditional computing which executes the entire model on the local mobile device, edge computing offloads the heavy computation tasks to an edge or cloud server that has more resources. Because deploying machine learning models on mobile hardware raises computational challenges, sending compressed data to an edge server to take care of the heavy computation can make more sense \cite{matsubara}. Development of efficient compression models tailored for edge computing, will enable improved resource utilization and enhanced performance in resource-constrained edge environments. 

\textbf{Edge Transmission.}
Due to the ever increasing amount of data transmitted from edge sensors, including satellite imagery, there is continuous research into optimizing how data is transferred and filtering which data is transferred from the edge.

Work being done to optimize data compression for transmission in mobile edge networks must consider the challenges and trade-offs associated with data transmission where limited network resources and varying communication conditions impact  efficiency \cite{ren}. Decisions about when and where to perform data compression to optimize the overall transmission process must be reflective of the current environment. 

%As network resources change, the amount of collected may vary which necessitates the modification of data in order to transmit at a sustained level before local resources run out.

There are also challenges of managing and storing large volumes of images generated by IoT devices. Because of the delay in downloading and storing large-scale image files, techniques are being developed to anticipate usage which will drive transmission strategies \cite{yin}. The efforts in this area seek to optimize the use of edge computing strategies to maximize storage efficiency while meeting real-time data access demands of IoT devices. 

One aspect of optimizing for transmission to and from the edge is saliency determination. For instance, in surveillance systems, being able to generate large amounts of data and perform real-time analysis to detect salient objects or events proves to be a challenge \cite{zhang}. The potential applications and benefits of on demand saliency determination encompass various industrial domains, such as security monitoring, anomaly detection, and object tracking. 

%The findings from their experiments contribute to the development of intelligent surveillance systems in industrial environments, improving situational awareness and enhancing the overall security and efficiency of industrial operations. 

%todo saliency determination from on board satellite

\textbf{Machine Learning.} Saliency driven compression can also optimize workflows for machine learning applications. Focusing on the information that will be most important to the downstream ML task can allow for reduced file sizes. Methods to choose compression schemes based on expected inference accuracy can allow end-users to adapt to changes in image input and model requirements \cite{3447878}. Work is also being done to optimize sensors based on computer vision pipelines that include compression \cite{Buckler_2017_ICCV}. In this effort a study was conducted as to the importance of different modules in an image processing pipeline and then a new sensor was proposed to replicate those stages, allowing them to be skipped in the pipeline. One of the most important stages found was that of gamma compression. Hardware optimizations have also been developed to help choose compression schemes for a desired reconstruction error threshold \cite{s140406247}. These onboard advisors can be adapted to different hardware constraints, including satellite based requirements, and are extensible to allow multiple predictive considerations into their planning.

\end{document}